\documentclass[11pt,a4paper]{article}
\usepackage[hyperref]{naaclhlt2019}
\usepackage{times}
\usepackage{latexsym}

\usepackage{url}

\usepackage{amsmath}
\usepackage{lipsum}
\usepackage{latexsym}
\usepackage[latin1]{inputenc}
\usepackage{amssymb}
\usepackage{graphicx}
\usepackage{listings}
\usepackage{color}
\usepackage{array}
\usepackage{setspace}
\usepackage{hyperref}
\usepackage{float}
\usepackage{multirow}
\usepackage{enumitem}

\newcommand{\bv}{\textbf{v}}

\definecolor{mygreen}{rgb}{0.114, 0.529, 0.106}

\DeclareMathOperator*{\argmax}{arg\,max}

\usepackage[skip=4pt]{caption}
\usepackage{natbib}
\usepackage{todonotes}
\usepackage{xcolor}
\usepackage{booktabs}
\usepackage{bbm}

\aclfinalcopy 

\title{Learning to Denoise Distantly-Labeled Data for Entity Typing}

\author{Yasumasa Onoe \and Greg Durrett\\
  Department of Computer Science \\
  The University of Texas at Austin \\
  {\tt\{yasumasa, gdurrett\}@cs.utexas.edu}}

\date{}

\begin{document}
\maketitle
\begin{abstract}
Distantly-labeled data can be used to scale up training of statistical models, but it is typically noisy and that noise can vary with the distant labeling technique. In this work, we propose a two-stage procedure for handling this type of data: denoise it with a learned model, then train our final model on clean and denoised distant data with standard supervised training. Our denoising approach consists of two parts. First, a filtering function discards examples from the distantly labeled data that are wholly unusable. Second, a relabeling function repairs noisy labels for the retained examples. Each of these components is a model trained on synthetically-noised examples generated from a small manually-labeled set.  We investigate this approach on the ultra-fine entity typing task of \citet{Eunsol_Choi_18}. Our baseline model is an extension of their model with pre-trained ELMo representations, which already achieves state-of-the-art performance. Adding distant data that has been denoised with our learned models gives further performance gains over this base model, outperforming models trained on raw distant data or heuristically-denoised distant data.

\end{abstract}


\section{Introduction}\label{intro}

With the rise of data-hungry neural network models, system designers have turned increasingly to unlabeled and weakly-labeled data in order to scale up model training. For information extraction tasks such as relation extraction and entity typing, distant supervision \cite{Mike_Mintz_09} is a powerful approach for adding more data, using a knowledge base \cite{delcorro-EtAl:2015:EMNLP,rabinovich-klein:2017:Short} or heuristics \cite{ratner16-data-programming,hancock2018} to automatically label instances. One can treat this data just like any other supervised data, but it is noisy; more effective approaches employ specialized probabilistic models \cite{Sebastian_Riedel10,ratner-snorkel}, capturing its interaction with other supervision \cite{Hai_Wang_18} or breaking down aspects of a task on which it is reliable \cite{Alexander_Ratner_18}. However, these approaches often require sophisticated probabilistic inference for training of the final model. Ideally, we want a technique that handles distant data just like supervised data, so we can treat our final model and its training procedure as black boxes.

This paper tackles the problem of exploiting weakly-labeled data in a structured setting with a two-stage denoising approach. We can view a distant instance's label as a noisy version of a true underlying label. We therefore learn a model to turn a noisy label into a more accurate label, then apply it to each distant example and add the resulting denoised examples to the supervised training set. Critically, the denoising model can condition on both the example and its noisy label, allowing it to fully leverage the noisy labels, the structure of the label space, and easily learnable correspondences between the instance and the label.


\begin{figure*}[t]
\centering
    \begin{minipage}[t]{0.45\textwidth}
    \centering
    \includegraphics[width=1.0\linewidth]{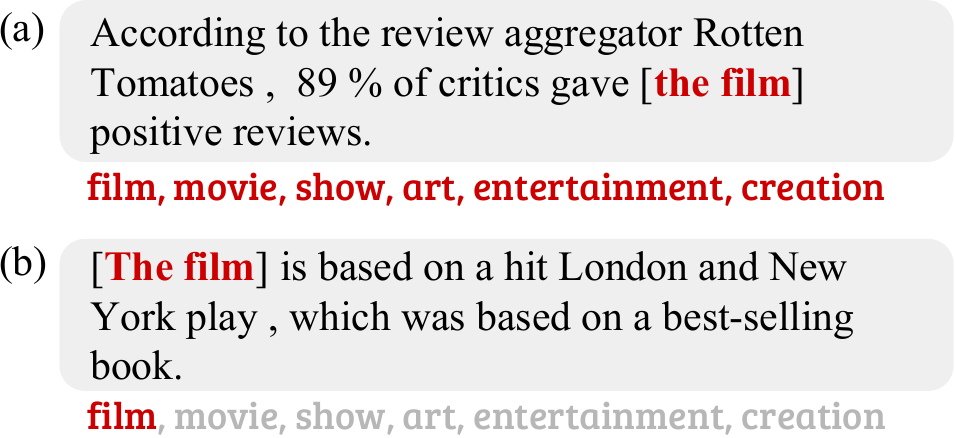}
    \end{minipage}
    \hspace{24pt}
    \begin{minipage}[t]{0.45\textwidth}
    \centering
    \includegraphics[width=1.0\linewidth]{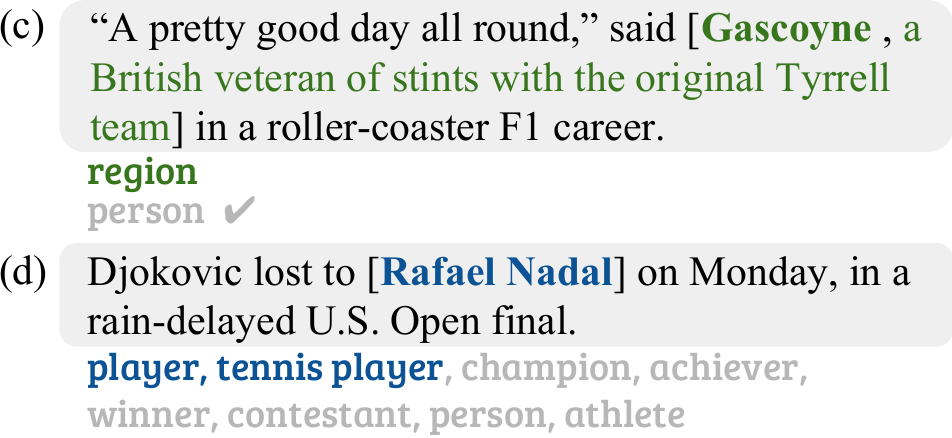}
    \end{minipage}
    \caption{Examples selected from the Ultra-Fine Entity Typing dataset of \citet{Eunsol_Choi_18}. (a) A manually-annotated example. (b) The head word heuristic functioning correctly but missing types in (a). (c) Entity linking providing the wrong types. (d) Entity linking providing correct but incomplete types.}
    \label{fig:ds-example}
\end{figure*}


Concretely, we implement our approach for the task of fine-grained entity typing, where a single entity may be assigned many labels. We learn two denoising functions: a \emph{relabeling} function takes an entity mention with a noisy set of types and returns a cleaner set of types, closer to what manually labeled data has. A \emph{filtering} function discards examples which are deemed too noisy to be useful. These functions are learned by taking manually-labeled training data, synthetically adding noise to it, and learning to denoise, similar to a conditional variant of a denoising autoencoder \cite{Pascal_Vincent_08}. Our denoising models embed both entities and labels to make their predictions, mirroring the structure of the final entity typing model itself.

We evaluate our model following \citet{Eunsol_Choi_18}. We chiefly focus on their ultra-fine entity typing scenario and use the same two distant supervision sources as them, based on entity linking and head words. On top of an adapted model from \citet{Eunsol_Choi_18} incorporating ELMo \cite{ELMO_18}, na\"{i}vely adding distant data actually hurts performance. However, when our learned denoising model is applied to the data, performance improves, and it improves more than heuristic denoising approaches tailored to this dataset. Our strongest denoising model gives a gain of 3 F$_1$ absolute over the ELMo baseline, and a 4.4 F$_1$ improvement over naive incorporation of distant data. This establishes a new state-of-the-art on the test set, outperforming concurrently published work \cite{Wenhan_Xiong_2019} and matching the performance of a BERT model \cite{BERT18} on this task. Finally, we show that denoising helps even when the label set is projected onto the OntoNotes label set \cite{HovyEtAl2006,Dan_Gillick_14}, outperforming the method of \citet{Eunsol_Choi_18} in that setting as well.


\section{Setup}\label{setup}
We consider the task of predicting a structured target $y$ associated with an input $x$. Suppose we have high-quality labeled data of $n$ (input, target) pairs $\mathcal{D} = \{\left(x^{(1)}, y^{(1)}\right), \dots, (x^{(n)}, y^{(n)})\}$, and noisily labeled data of $n'$ (input, target) pairs $\mathcal{D}' = \{(x^{(1)}, y^{(1)}_{\text{noisy}}), \dots, (x^{(n')}, y^{(n')}_{\text{noisy}})\}$. For our tasks, $\mathcal{D}$ is collected through manual annotation and $\mathcal{D}'$ is collected by distant supervision. 
We use two models to denoise data from $\mathcal{D}'$: a filtering function $f$ disposes of unusable data (e.g., mislabeled examples) and a relabeling function $g$ transforms the noisy target labels $y_{\text{noisy}}$ to look more like true labels. This transformation improves the noisy data so that we can use it to $\mathcal{D}$ without introducing damaging amounts of noise. In the second stage, a classification model is trained on the augmented data ($\mathcal{D}$ combined with denoised $\mathcal{D}'$) and predicts $y$ given $x$ in the inference phase.

\subsection{Case Study: Ultra-Fine Entity Typing}

The primary task we address here is the fine-grained entity typing task of \citet{Eunsol_Choi_18}. Instances in the corpus are assigned types from a vocabulary of more than 10,000 types, which are divided into three classes: $9$ \textit{general} types, $121$ \textit{fine-grained} types, and $10,201$ \textit{ultra-fine} types. This dataset consists of 6K manually annotated examples and approximately 25M distantly-labeled examples. 5M examples are collected using entity linking (EL) to link mentions to Wikipedia and gather types from information on the linked pages. 20M examples (HEAD) are generated by extracting nominal head words from raw text and treating these as singular type labels.


\begin{figure*}[t]
\centering
    \centering
    \includegraphics[width=1.0\linewidth]{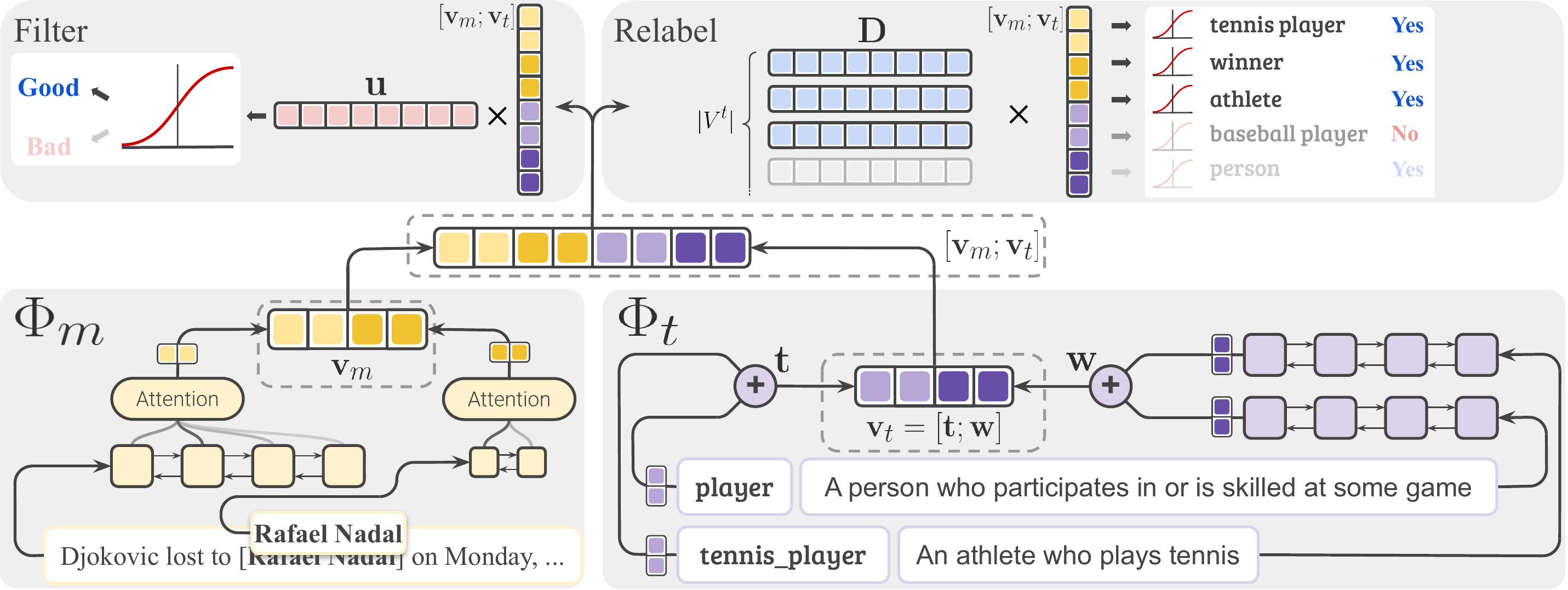}
    \caption{Denoising models. The \emph{Filter} model predicts whether the example should be kept at all; if it is kept, the \emph{Relabel} model attempts to automatically expand the label set. $\Phi_m$ is a mention encoder, which can be a state-of-the-art entity typing model. $\Phi_t$ encodes noisy types from distant supervision.}
    \label{fig:high-level}
\end{figure*}


Figure~\ref{fig:ds-example} shows examples from these datasets which illustrate the challenges in automatic annotation using distant supervision. The manually-annotated example in (a) shows how numerous the gold-standard labeled types are. By contrast, the HEAD example (b) shows that simply treating the head word as the type label, while correct in this case, misses many valid types, including more general types. The EL example (c) is incorrectly annotated as \texttt{region}, whereas the correct coarse type is actually \texttt{person}. This error is characteristic of entity linking-based distant supervision since identifying the correct link is a challenging problem in and of itself \cite{MilneWitten2008}: in this case, \emph{Gascoyne} is also the name of a region in Western Australia. The EL example in (d) has reasonable types; however, human annotators could choose more types (grayed out) to describe the mention more precisely. The average number of types annotated by humans is $5.4$ per example while the two distant supervision techniques combined yields $1.5$ types per example on average.

In summary, distant supervision can (1) produce completely incorrect types, and (2) systematically miss certain types.


\section{Denoising Model}\label{cdae}

To handle the noisy data, we propose to learn a denoising model as shown in Figure~\ref{fig:high-level}. This denoising model consists of filtering and relabeling functions to discard and relabel examples, respectively; these rely on a shared mention encoder and type encoder, which we describe in the following sections. The filtering function is a binary classifier that takes these encoded representations and predicts whether the example is good or bad. The relabeling function predicts a new set of labels for the given example.

We learn these functions in a supervised fashion. Training data for each is created through synthetic noising processes applied to the manually-labeled data, as described in Sections~\ref{filter} and \ref{relabel}.

For the entity typing task, each example $(x,y)$ takes the form $((s,m),t)$, where $s$ is the sentence, $m$ is the mention span, and $t$ is the set of types (either clean or noisy).

\subsection{Mention Encoder}

This encoder is a function $\Phi_m(s,m)$ which maps a sentence $s$ and mention $m$ to a real-valued vector $\bv_m$. This allows the filtering and relabeling function to recognize inconsistencies between the given example and the provided types. Note that these inputs $s$ and $m$ are the same as the inputs for the supervised version of this task; we can therefore share an encoder architecture between our denoising model and our final typing model. We use an encoder following \citet{Eunsol_Choi_18} with a few key differences, which are described in Section~\ref{model_details}.

\subsection{Type Encoder}

The second component of our model is a module which produces a vector $\bv_t = \Phi_t(t)$. This is an encoder of an unordered bag of types. Our basic type encoder uses trainable vectors as embeddings for each type and combines these with summing. That is, the noisy types $t_1, \dots, t_m$ are embedded into type vectors $\{\mathbf{t}_1, \dots, \mathbf{t}_m\}$.
The final embedding of the type set $\mathbf{t} = \sum_{j} \mathbf{t}_j $. 

\paragraph{Type Definition Encoder} Using trainable type embeddings exposes the denoising model to potential data sparsity issues, as some types appear only a few or zero times in the training data.  Therefore, we also assign each type a vector based on its definition in WordNet \cite{wordnet_1995}. Even low-frequent types are therefore assigned a plausible embedding.\footnote{We found this technique to be more effective than using pretrained vectors from GloVe or ELMo. It gave small improvements on an intrinsic evaluation over not incorporating it; results are omitted due to space constraints.}

Let $w^j_i$ denote the $i$th word of the $j$th type's most common WordNet definition. Each $w^j_i$ is embedded using GloVe \cite{glove_14}. The resulting word embedding vectors $\mathbf{w}_i^j$ are fed into a bi-LSTM \cite{lstm, bilstm}, and a concatenation of the last hidden states in both directions is used as the definition representation $\mathbf{w}^j$. The final representation of the definitions is the sum over these vectors for each type: $\mathbf{w} = \sum_{k} \mathbf{w}^k$.  

Our final $\mathbf{v}_t = [\mathbf{t}; \mathbf{w}]$, the concatenation of the type and definition embedding vectors.

\vspace{6pt}

\subsection{Filtering Function}\label{filter}

The filtering function $f$ is a binary classifier designed to detect examples that are completely mislabeled. Formally, $f$ is a function mapping a labeled example $(s,m,t)$ to a binary indicator $z$ of whether this example should be discarded or not.

In the forward computation, the feature vectors $\mathbf{v}_m$ and $\mathbf{v}_t$ are computed using the mention and type encoders. The model prediction is defined as $P(\textrm{error}) = \sigma \left(\mathbf{u}^\top \text{Highway}\left(\left[\mathbf{v}_m; \mathbf{v}_t \right]\right)\right)$, where $\sigma$ is a sigmoid function, $\mathbf{u}$ is a parameter vector, and $\text{Highway}(\cdot)$ is a 1-layer highway network \cite{highway15}. We can apply $f$ to each distant pair in our distant dataset $\mathcal{D}'$ and discard any example predicted to be erroneous ($P(\textrm{error}) > 0.5$).

\vspace{-4pt}
\paragraph{Training data} We do not know a priori which examples in the distant data should be discarded, and labeling these is expensive. We therefore construct synthetic training data $\mathcal{D}_{\text{error}}$ for $f$ based on the manually labeled data $\mathcal{D}$. For 30\% of the examples in $\mathcal{D}$, we replace the gold types for that example with \emph{non-overlapping} types taken from another example. The intuition for this procedure follows Figure~\ref{fig:ds-example}: we want to learn to detect examples in the distant data like \emph{Gascoyne} where heuristics like entity resolution have misfired and given a totally wrong label set.

Formally, for each selected example $((s,m), t)$, we repeatedly draw another example $((s',m'), t')$ from $\mathcal{D}$ until we find $t'_{\text{error}}$ that does not have any common types with $t$. We then create a positive training example $((s,m,t'_{\text{error}}), z = 1)$. We create a negative training example $((s,m,t), z = 0)$ using the remaining $70\%$ of examples. $f$ is trained on $\mathcal{D}_{\text{error}}$ using binary cross-entropy loss.

\subsection{Relabeling Function}\label{relabel}

The relabeling function $g$ is designed to repair examples that make it through the filter but which still have errors in their type sets, such as missing types as shown in Figure~\ref{fig:ds-example}b and \ref{fig:ds-example}d. $g$ is a function from a labeled example $(s,m,t)$ to an improved type set $\tilde{t}$ for the example.

Our model computes feature vectors $\mathbf{v}_m$ and $\mathbf{v}_t$ by the same procedure as the filtering function $f$. The decoder is a linear layer with parameters $\mathbf{D} \in \mathbb{R}^{|V^{\text{t}}| \times (d_m + d_t)}$. We compute $\mathbf{e} = \sigma \left(\mathbf{D} \left[\mathbf{v}_m; \mathbf{v}_t \right] \right)$, where $\sigma$ is an element-wise sigmoid operation designed to give binary probabilities for each type.

Once $g$ is trained, we make a prediction $\tilde{t}$ for each $(s,m,t) \in \mathcal{D}'$ and replace $t$ by $\tilde{t}$ to create the denoised data $\mathcal{D}'_{\text{denoise}} = \{(s,m,\tilde{t}), \dots \}$. For the final prediction, we choose all types $t_\ell$ where $e_\ell > 0.5$, requiring at least two types to be present or else we discard the example.

\vspace{-4pt}
\paragraph{Training data} We train the relabeling function $g$ on another synthetically-noised dataset $\mathcal{D}_{\text{drop}}$ generated from the manually-labeled data $\mathcal{D}$. To mimic the type distribution of the distantly-labeled examples, we take each example $(s,m,t)$ and randomly drop each type with a fixed rate $0.7$ independent of other types to produce a new type set $t'$. We perform this process for all examples in $\mathcal{D}$ and create a noised training set $\mathcal{D}_{\text{drop}}$, where a single training example is $((s,m,t'), t)$. $g$ is trained on $\mathcal{D'}_{\text{drop}}$ with a binary classification loss function over types used in \citet{Eunsol_Choi_18}, described in the next section.

One can think of $g$ as a type of denoising autoencoder \cite{Pascal_Vincent_08} whose reconstructed types $\tilde{t}$ are conditioned on $\mathbf{v}$ as well as $t$.


\section{Typing Model}\label{model_details}

In this section, we define the sentence and mention encoder $\Phi_m$, which is use both in the denoising model as well as in the final prediction task. We extend previous attention-based models for this task \cite{Sonse_Shimaoka_17,Eunsol_Choi_18}. At a high level, we have an instance encoder $\Phi_m$ that returns a vector $\bv_m \in \mathbb{R}^{d_\Phi}$, then multiply the output of this encoding by a matrix and apply a sigmoid to get a binary prediction for each type as a probability of that type applying.

\begin{figure}[t]
\centering
    \centering
    \includegraphics[width=1.0\linewidth]{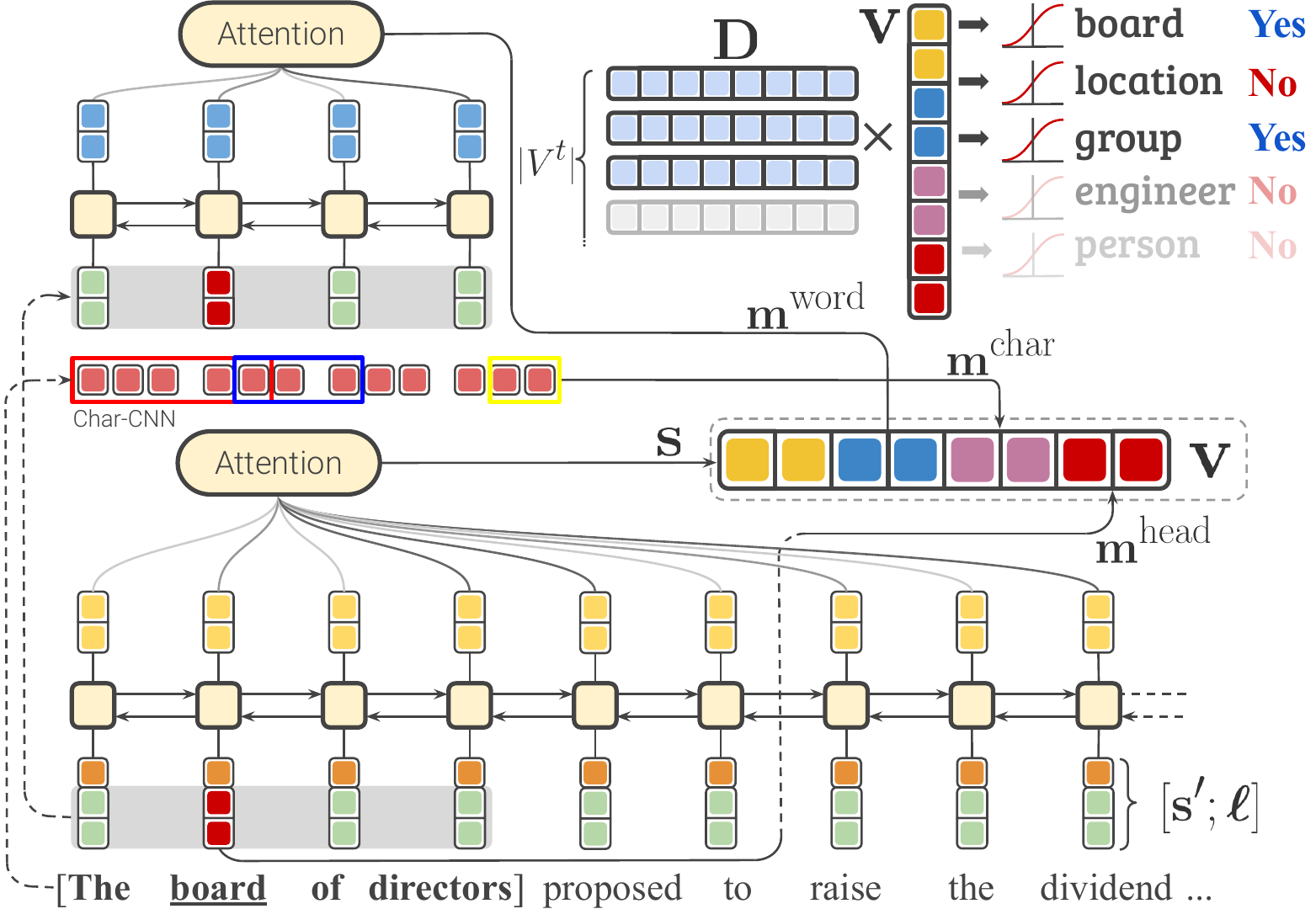}
    \caption{Sentence and mention encoder used to predict types. We compute attention over LSTM encodings of the sentence and mention, as well as using character-level and head-word representations to capture additional mention properties. These combine to form an encoding which is used to predict types.}
    \label{fig:model}
\end{figure}

Figure~\ref{fig:model} outlines the overall architecture of our typing model. The encoder $\Phi_m$ consists of four vectors: a sentence representation $\mathbf{s}$, a word-level mention representation $\mathbf{m^{\text{word}}}$, a character-level mention representation $\mathbf{m}^{\text{char}}$, and a headword mention vector $\mathbf{m}^{\text{head}}$. The first three of these were employed by \citet{Eunsol_Choi_18}. We have modified the mention encoder with an additional bi-LSTM to better encode long mentions, and additionally used the headword embedding directly in order to focus on the most critical word. These pieces use pretrained contextualized word embeddings (ELMo) \cite{ELMO_18} as input. 

\paragraph{Pretrained Embeddings} 
Tokens in the sentence $s$ are converted into contextualized word vectors using ELMo; let $\mathbf{s}'_i \in \mathbb{R}^{d_{ELMo}}$ denote the embedding of the $i$th word. As suggested in \citet{ELMO_18}, we learn task specific parameters $\gamma^{\text{task}} \in \mathbb{R}$ and $\mathbf{s}^{\text{task}} \in \mathbb{R}^3$ governing these embeddings. We do not fine-tune the parameters of the ELMo LSTMs themselves.


\vspace{-4pt}
\paragraph{Sentence Encoder}  Following \citet{Eunsol_Choi_18}, we concatenate the $m$th word vector $\mathbf{s}_m$ in the sentence with a corresponding location embedding $\boldsymbol\ell_m \in \mathbb{R}^{d_{\text{loc}}}$. Each word is assigned one of four location tokens, based on whether (1) the word is in the left context, (2) the word is the first word of the mention span, (3) the word is in the mention span (but not first), and (4) the word is in the right context. The input vectors $[\mathbf{s}'; \boldsymbol\ell]$ are fed into a bi-LSTM encoder, with hidden dimension is $d_{\text{hid}}$, followed by a span attention layer \cite{Kenton_Lee17,Eunsol_Choi_18}: $\mathbf{s} = \text{Attention}(\text{bi-LSTM}([\mathbf{s}'; \mathbf{l}]))$, where $\mathbf{s}$ is the final representation of the sentence $s$.

\vspace{-4pt}
\paragraph{Mention Encoder} To obtain a mention representation, we use both word and character information. For the word-level representation, the mention's contextualized word vectors $\mathbf{m}'$ are fed into a bi-LSTM with hidden dimension is $d_{\text{hid}}$. The concatenated hidden states of both directions are summed by a span attention layer to form the word-level mention representation: $\mathbf{m}^{\text{word}} = \text{Attention}(\text{bi-LSTM}(\mathbf{m}'))$.

Second, a character-level representation is computed for the mention. Each character is embedded and then a 1-D convolution \cite{nlpfromscratch11} is applied over the characters of the mention. This gives a character vector $\mathbf{m}^{\text{char}}$.

Finally, we take the contextualized word vector of the headword  $\mathbf{m}^{\text{head}}$ as a third component of our representation.  This can be seen as a residual connection \cite{resnet16} specific to the mention head word.  We find the headwords in the mention spans by parsing those spans in isolation using the spaCy dependency parser \cite{spacy_15}. Empirically, we found this to be useful on long spans, when the span attention would often focus on incorrect tokens.

The final representation of the input $x$ is a concatenation of the sentence, the word- \& character-level mention, and the mention headword representations, $\bv = \left[\mathbf{s}; \mathbf{m}^{\text{word}}; \mathbf{m}^{\text{char}}; \mathbf{m}^{\text{head}} \right] \in \mathbb{R}^{d_\Phi}$.

\vspace{-4pt}
\paragraph{Decoder}  We treat each label prediction as an independent binary classification problem. Thus, we compute a score for each type in the type vocabulary $V^t$. 
Similar to the decoder of the relabeling function $g$, we compute $\mathbf{e} = \sigma \left( \mathbf{E} \bv \right)$, where $\mathbf{E} \in \mathbb{R}^{|V^{\text{t}}| \times d_\Phi}$ and  $\mathbf{e} \in \mathbb{R}^{|V^{\text{t}}|}$. For the final prediction, we choose all types $t_\ell$ where $e_\ell > 0.5$. If none of $e_\ell$ is greater than $0.5$, we choose $t_\ell = \argmax \mathbf{e}$ (the single most probable type).

\vspace{-4pt}
\paragraph{Loss Function} We use the same loss function as \citet{Eunsol_Choi_18} for training. This loss partitions the labels in general, fine, and ultra-fine classes, and only treats an instance as an example for types of the class in question if it contains a label for that class. More precisely:
\vspace{-6pt}
\begin{equation}
\label{loss1}
\begin{aligned}
\mathcal{L} = &\mathcal{L}_{\text{general}} \mathbbm{1}_{\text{general}}(t) + \mathcal{L}_{\text{fine}} \mathbbm{1}_{\text{fine}}(t) \\
&+ \mathcal{L}_{\text{ultra-fine}} \mathbbm{1}_{\text{ultra-fine}}(t),
\end{aligned}
\end{equation}

where $\mathcal{L}_{\dots}$ is a loss function for a specific type class: general, fine-grained, or ultra-fine, and $\mathbbm{1}_{\dots}(t)$ is an indicator function that is active when one of the types $t$ is in the type class. Each $\mathcal{L}_{\dots}$ is a sum of binary cross-entropy losses over all types in that category. That is, the typing problem is viewed as independent classification for each type.

Note that this loss function already partially repairs the noise in distant examples from missing labels: for example, it means that examples from HEAD do not count as negative examples for general types when these are not present. However, we show in the next section that this is not sufficient for denoising.

\vspace{-4pt}
\paragraph{Implementation Details} The settings of hyperparameters in our model largely follows \citet{Eunsol_Choi_18} and recommendations for using the pretrained ELMo-Small model.\footnote{\url{https://allennlp.org/elmo}} The word embedding size $d_{\text{ELMo}}$ is $1024$. The type embedding size and the type definition embedding size are set to $1024$. For most of other model hyperparameters, we use the same settings as \citet{Eunsol_Choi_18}: $d_{\text{loc}} = 50$, $d_{\text{hid}} = 100$, $d_{\text{char}} = 100$.
The number of filters in the 1-d convolutional layer is $50$. Dropout is applied with $p = 0.2$ for the pretrained embeddings, and $p = 0.5$ for the mention representations. We limit sentences to 50 words and mention spans to 20 words for computational reasons. The character CNN input is limited to $25$ characters; most mentions are short, so this still captures subword information in most cases. The batch size is set to $100$. For all experiments, we use the Adam optimizer \cite{Kingma_14}. The initial learning rate is set to 2e-03. We implement all models\footnote{The code for experiments is available at \url{https://github.com/yasumasaonoe/DenoiseET}} using PyTorch. To use ELMo, we consult the AllenNLP source code.


\section{Experiments}\label{experiments}

\paragraph{Ultra-Fine Entity Typing}
We evaluate our approach on the ultra-fine entity typing dataset from \citet{Eunsol_Choi_18}.  The 6K manually-annotated English examples are equally split into the training, development, and test examples by the authors of the dataset. We generate synthetically-noised data, $\mathcal{D}_{\text{error}}$ and $\mathcal{D}_{\text{drop}}$, using the 2K training set to train the filtering and relabeling functions, $f$ and $h$. We randomly select 1M EL and 1M HEAD examples and use them as the noisy data $\mathcal{D}'$. Our augmented training data is a combination of the manually-annotated data $\mathcal{D}$ and $\mathcal{D}'_{\text{denoised}}$. 

\vspace{-4pt}
\paragraph{OntoNotes}
In addition, we investigate if denoising leads to better performance on another dataset. We use the English OntoNotes dataset \cite{Dan_Gillick_14}, which is a widely used benchmark for fine-grained entity typing systems. The original training, development, and test splits contain 250K, 2K, and 9K examples respectively. \citet{Eunsol_Choi_18} created an augmented training set that has 3.4M examples. We also construct our own augmented training sets with/without denoising using our noisy data $\mathcal{D}'$, using the same label mapping from ultra-fine types to OntoNotes types described in \citet{Eunsol_Choi_18}.

\subsection{Ultra-Fine Typing Results}

We first compare the performance of our approach to several benchmark systems, then break down the improvements in more detail. We use the model architecture described in Section~\ref{model_details} and train it on the different amounts of data: manually labeled only, naive augmentation (adding in the raw distant data), and denoised augmentation. We compare our model to \citet{Eunsol_Choi_18} as well as to BERT \cite{BERT18}, which we fine-tuned for this task. We adapt our task to BERT by forming an input sequence "\texttt{[CLS] sentence [SEP] mention [SEP]}" and assign the segment embedding A to the sentence and B to the mention span.\footnote{We investigated several approaches, including taking the head word piece from the last layer and using that for classification (more closely analogous to what \citet{BERT18} did for NER), but found this one to work best.} Then, we take the output vector at the position of the \texttt{[CLS]} token (i.e., the first token) as the feature vector $\bv$, analogous to the usage for sentence pair classification tasks. The BERT model is fine-tuned on the 2K manually annotated examples. We use the pretrained BERT-Base, uncased model\footnote{\url{https://github.com/google-research/bert}} with a step size of 2e-05 and batch size 32.

\vspace{-4pt}
\paragraph{Results} Table~\ref{tab:ultra-fine-dev-breakdown} compares the performance of these systems on the development set. Our model with no augmentation already matches the system of \citet{Eunsol_Choi_18} with augmentation, and incorporating ELMo gives further gains on both precision and recall. On top of this model, adding the distantly-annotated data lowers the performance; the loss function-based approach of \cite{Eunsol_Choi_18} does not sufficiently mitigate the noise in this data. However, denoising makes the distantly-annotated data useful, improving recall by a substantial margin especially in the general class. A possible reason for this is that the relabeling function tends to add more general types given finer types. BERT performs similarly to ELMo with denoised distant data. As can be seen in the performance breakdown, BERT gains from improvements in recall in the fine class.

\renewcommand{\arraystretch}{1}
\begin{table*}[t]
	\centering
	\small
	\setlength{\tabcolsep}{4pt}
	\begin{tabular}{l c c c c c c c c c c c c}
		\toprule
		\multicolumn{1}{c}{} & \multicolumn{3}{c}{Total} & \multicolumn{3}{c}{General} & \multicolumn{3}{c}{Fine} & \multicolumn{3}{c}{Ultra-Fine} \\
	    \cmidrule(r){2-4}  \cmidrule(r){5-7} \cmidrule(r){8-10} \cmidrule(r){11-13}
		\multicolumn{1}{c}{Model}
		 & P & R & F1  & P & R & F1 & P & R & F1 & P & R & F1\\
		\midrule
		Ours + GloVe w/o augmentation & 46.4 & 23.3 & 31.0  & 57.7 & 65.5 & 61.4 & 41.3 & 31.3 & 35.6 & 42.4 & 9.2 & 15.1\\
		Ours + ELMo w/o augmentation & \textbf{55.6} & 28.1 & 37.3  & \textbf{69.3} & 77.3 & 73.0 & \textbf{47.9} & 35.4 & 40.7 & \textbf{48.9} & 12.6 & 20.0\\
		Ours + ELMo w augmentation & 55.2 & 26.4 & 35.7  & 69.4 & 72.0 & 70.7 & 46.6 & 38.5 & 42.2 & 48.7 & 10.3 & 17.1\\
		Ours + ELMo w augmentation & 50.7 & \textbf{33.1} & \textbf{40.1}  & 66.9 & \textbf{80.7} & 73.2 & 41.7 & 46.2 & 43.8 & 45.6 & \textbf{17.4} & \textbf{25.2}\\
		\hspace{48pt} + filter \& relabel &  &  &  &  &  &  &  &  &  &  &  & \\
		BERT-Base, Uncased & 51.6 & 32.8 & \textbf{40.1}  & 67.4 & 80.6 & \textbf{73.4} & 41.6 & \textbf{54.7} & \textbf{47.3} & 46.3 & 15.6 & 23.4\\
		\midrule
		\citet{Eunsol_Choi_18} w augmentation &  48.1 & 23.2 & 31.3 & 60.3 & 61.6 & 61.0 & 40.4 & 38.4 & 39.4 & 42.8 & 8.8 & 14.6\\
		\bottomrule 
	\end{tabular}
	\caption{Macro-averaged P/R/F1 on the dev set for the entity typing task of \citet{Eunsol_Choi_18} comparing various systems. ELMo gives a substantial improvement over baselines. Over an ELMo-equipped model, data augmentation using the method of \citet{Eunsol_Choi_18} gives no benefit. However, our denoising technique allow us to effectively incorporate distant data, matching the results of a BERT model on this task \cite{BERT18}.} \label{tab:ultra-fine-dev-breakdown}
\end{table*}

Table~\ref{tab:ultra-fine-test} shows the performance of all settings on the test set, with the same trend as the performance on the development set. Our approach outperforms the concurrently-published \citet{Wenhan_Xiong_2019}; however, that work does not use ELMo. Their improved model could be used for both denoising as well as prediction in our setting, and we believe this would stack with our approach.

\vspace{-4pt}
\paragraph{Usage of Pretrained Representations} Our model with ELMo trained on denoised data matches the performance of the BERT model. We experimented with incorporating distant data (raw and denoised) in BERT, but the fragility of BERT made it hard to incorporate: training for longer generally caused performance to go down after a while, so the model cannot exploit large external data as effectively. \citet{BERT18} prescribe training with a small batch size and very specific step sizes, and we found the model very sensitive to these hyperparameters, with only 2e-05 giving strong results. The ELMo paradigm of incorporating these as features is much more flexible and modular in this setting. Finally, we note that our approach could use BERT for denoising as well, but this did not work better than our current approach. Adapting BERT to leverage distant data effectively is left for future work.

\subsubsection{Comparing Denoising Models}

We now explicitly compare our denoising approach to several baselines. For each denoising method, we create the denoised EL, HEAD, and EL\:\&\:HEAD dataset and investigate performance on these datasets. Any denoised dataset is combined with the 2K manually-annotated examples and used to train the final model.

\vspace{-4pt}
\paragraph{Heuristic Baselines} These heuristics target the same factors as our filtering and relabeling functions in a non-learned way.
\begin{description}[align=left, font=\sc, topsep=2pt, labelsep=\fontdimen2\font, leftmargin=0pt]
\setlength\itemsep{-0.1em}
\item [Synonyms and Hypernyms] \hspace{2pt} For each type observed in the distant data, we add its synonyms and hypernyms using WordNet \cite{wordnet_1995}. This is motivated by the data construction process in  \citet{Eunsol_Choi_18}.
\item [Common Type Pairs]  \hspace{2pt} We use type pair statistics in the manually labeled training data. For each base type that we observe in a distant example, we add any type which is seen more than 90\% of the time the base type occurs. For instance, the type \texttt{art} is given at least 90\% of the times the \texttt{film} type is present, so we automatically add \texttt{art} whenever \texttt{film} is observed.  
\item [Overlap] \hspace{2pt} We train a model on the manually-labeled data only, then run it on the distantly-labeled data. If there is an intersection between the noisy types $t$ and the predicted type $\hat t$, we combine them and use as the expanded type $\tilde t$. Inspired by tri-training \cite{tri05}, this approach adds ``obvious'' types but avoids doing so in cases where the model has likely made an error. 
\end{description}

\renewcommand{\arraystretch}{1}
\begin{table}[t]
	\centering
	\small
	\setlength{\tabcolsep}{4pt}
	\begin{tabular}{l  c  c  c }
		\toprule
		\multicolumn{1}{c }{Model}
		 & P & R & F1\\
		\midrule
		Ours + GloVe w/o augmentation & 47.6 & 23.3 & 31.3  \\
		Ours + ELMo w/o augmentation & \textbf{55.8} & 27.7 & 37.0 \\
		Ours + ELMo w augmentation & 55.5 & 26.3 & 35.7 \\
		Ours + ELMo w augmentation & 51.5 & \textbf{33.0} & \textbf{40.2} \\
		\hspace{48pt} + filter \& relabel & &  \\
		BERT-Base, Uncased  & 51.6 & \textbf{33.0} & \textbf{40.2} \\
		\midrule
		\citet{Eunsol_Choi_18} w augmentation & 47.1 & 24.2 & 32.0  \\
		\textsc{LabelGCN} \cite{Wenhan_Xiong_2019} & 50.3 & 29.2 & 36.9\\
		\bottomrule 
	\end{tabular}
	\caption{Macro-averaged P/R/F1 on the test set for the entity typing task of \citet{Eunsol_Choi_18}. Our denoising approach gives substantial gains over naive augmentation and matches the performance of a BERT model.}
	\label{tab:ultra-fine-test}
\end{table}

\renewcommand{\arraystretch}{1}
\begin{table*}[t]
	\centering
	\small
	\setlength{\tabcolsep}{4pt}
	\begin{tabular}{l  l  c  c  c  c  c  c  c  c  c  c  c  c}
		\toprule
		\multicolumn{1}{c}{} & \multicolumn{1}{c}{} & \multicolumn{1}{c}{} & \multicolumn{3}{c}{EL\:\&\:HEAD} & \multicolumn{1}{c}{} & \multicolumn{3}{c}{EL} & \multicolumn{1}{c}{} & \multicolumn{3}{c}{HEAD}\\
	    \cmidrule(r){4-6}  \cmidrule(r){8-10} \cmidrule(r){12-14} 
		\multicolumn{1}{c}{Type} & \multicolumn{1}{c}{Denoising Method} & \multicolumn{1}{c}{} & P & R & F1 & & P & R & F1 & & P & R & F1\\
		\midrule
		\multirow{1}{*}{\text{}} & \textsc{Raw Data} & & \textbf{55.2} & 26.4 & 35.7 & & 52.3 & 26.1 & 34.8 & & \textbf{52.8} & 28.4 & 36.9\\
		\multirow{1}{*}{\text{Heuristic Baselines}} & \textsc{Synonyms\:\&\:Hypernyms} & & 43.0 & 30.0 & 35.3 & & 47.5 & 26.3 & 33.9 & & 44.8 & 31.7 & 37.1\\
		& \textsc{Pair} & & 50.2 & 29.0 & 36.8 & & 49.6 & 27.0 & 35.0 & & 50.6 & 31.2 & 38.6\\
		\vspace{4pt}
		& \textsc{Overlap} & & 50.0 & 32.3 & 39.2 & &  49.5 & \textbf{30.8} & 38.0 & & 50.6 & 31.4 & 38.7\\
		\multirow{1}{*}{\text{Proposed Approach}} & \textsc{Filter} & & 53.1 & 28.2 & 36.8 & & 51.9 & 26.5 & 35.1 & & 51.2 & 31.2 & 38.7\\
		& \textsc{Relabel} & & 52.1 & 32.2 & 39.8 & & 50.2 & 31.4 & 38.6 & & 50.2 & 31.8 & 38.9\\
		& \textsc{Filter\:\&\:Relabel} & & 50.7 & \textbf{33.1} & \textbf{40.1} & & \textbf{52.7} & 30.5 & \textbf{38.7} & & 50.7 & \textbf{32.1} & \textbf{39.3}\\
		\midrule
		& \citet{Eunsol_Choi_18} & &  48.1 & 23.2 & 31.3 & & 50.3 & 19.6 & 28.2 & & 48.4 & 22.3 & 30.6\\
		\bottomrule 
	\end{tabular}
	\caption{Macro-averaged P/R/F1 on the dev set for the entity typing task of \citet{Eunsol_Choi_18} with various types of augmentation added. The customized loss from \citet{Eunsol_Choi_18} actually causes a decrease in performance from adding any of the datasets. Heuristics can improve incorporation of this data: a relabeling heuristic (Pair) helps on HEAD and a filtering heuristic (Overlap) is helpful in both settings. However, our trainable filtering and relabeling models outperform both of these techniques.}\label{tab:ultra-fine-dev-denoise}
\end{table*}

\vspace{-4pt}
\paragraph{Results}  Table~\ref{tab:ultra-fine-dev-denoise} compares the results on the development set. We report the performance on each of the EL\:\&\:HEAD, EL, and HEAD dataset. On top of the baseline \textsc{Original}, adding synonyms and hypernyms by consulting external knowledge does not improve the performance. Expanding labels with the \textsc{Pair} technique results in small gains over \textsc{Original}. \textsc{Overlap} is the most effective heuristic technique.  This simple filtering and expansion heuristic improves recall on EL. \textsc{Filter}, our model-based example selector, gives similar improvements to \textsc{Pair} and \textsc{Overlap} on the HEAD setting, where filtering noisy data appears to be somewhat important.\footnote{One possible reason for this is identifying stray word senses; \emph{film} can refer to the physical photosensitive object, among other things.} \textsc{Relabel} and \textsc{Overlap} both improve performance on both EL and HEAD while other methods do poorly on EL. Combining the two model-based denoising techniques, \textsc{Filter\:\&\:Relabel} outperforms all the baselines.

\subsection{OntoNotes Results}\label{dis-onto}

We compare our different augmentation schemes for deriving data for the OntoNotes standard as well. Table~\ref{tab:onto-test} lists the results on the OntoNotes test set following the adaptation setting of \citet{Eunsol_Choi_18}. Even on this dataset, denoising significantly improves over naive incorporation of distant data, showing that the denoising approach is not just learning quirks of the ultra-fine dataset. Our augmented set is constructed from 2M seed examples while \citet{Eunsol_Choi_18} have a more complex procedure for deriving augmented data from 25M examples. Ours (total size of 2.1M) is on par with their larger data (total size of 3.4M), despite having 40\% fewer examples. In this setting, BERT still performs well but not as well as our model with augmented training data.

One source of our improvements from data augmentation comes from additional data that is able to be used because \emph{some} OntoNotes type can be derived. This is due to denoising doing a better job of providing correct general types. In the EL setting, this yields 730k usable examples out of 1M (vs 540K for no denoising), and in HEAD, 640K out of 1M (vs. 73K).

\renewcommand{\arraystretch}{1}
\begin{table}[t]
	\centering
	\vspace{-18pt}
	\small
	\setlength{\tabcolsep}{4pt}
	\begin{tabular}{lccc}
		\toprule
		\multicolumn{1}{c}{Model}
		 &{Acc.} & {Ma-F1}& {Mi-F1} \\
		\midrule
		Ours + ELMo w/o augmentation & 42.7 & 72.7  & 66.7\\
		Ours + ELMo w augmentation & 59.3 & 76.5 & 70.7\\
		Ours + ELMo w augmentation & 63.9 & \textbf{84.5} & 78.9\\
		\hspace{48pt} + filter \& relabel  & & \\
		Ours + ELMo w augmentation & \textbf{64.9} & \textbf{84.5} & \textbf{79.2}\\
		\hspace{40pt} by \citet{Eunsol_Choi_18}  &  &  \\
		BERT-Base, Uncased  & 51.8 & 76.6 & 69.1 \\
		\midrule
		\citet{Sonse_Shimaoka_17} & 51.7 & 70.9 & 64.9\\
	    AFET \cite{Xiang_Ren_16a} & 55.1 & 71.1 & 64.7 \\
	    PLE \cite{Xiang_Ren_16b} & 57.2 & 71.5 & 66.1 \\
		\citet{Eunsol_Choi_18} & 59.5 & 76.8 & 71.8 \\
		\textsc{LabelGCN} \cite{Wenhan_Xiong_2019} & 59.6 & 77.8 & 72.2\\
		\bottomrule 
	\end{tabular}
	\caption{Test results on OntoNotes. Denoising helps substantially even in this reduced setting. Using fewer distant examples, we nearly match the performance using the data from \citet{Eunsol_Choi_18} (see text).}
	\label{tab:onto-test}
\end{table}

\subsection{Analysis of Denoised Labels}

To understand what our denoising approach does to the distant data, we analyze the behavior of our filtering and relabeling functions. Table~\ref{tab:relabel} reports the average numbers of types added/deleted by the relabeling function and the ratio of examples discarded by the filtering function.

Overall, the relabeling function tends to add more and delete fewer number of types. The HEAD examples have more general types added than the EL examples since the noisy HEAD labels are typically finer. Fine-grained types are added to both EL and HEAD examples less frequently. Ultra-fine examples are frequently added to both datasets, with more added to EL; the noisy EL labels are mostly extracted from Wikipedia definitions, so those labels often do not include ultra-fine types. The filtering function discards similar numbers of examples for the EL and HEAD data: $9.4\%$ and $10\%$ respectively.

\begin{figure}[t]
    \centering
    \includegraphics[width=1.0\linewidth]{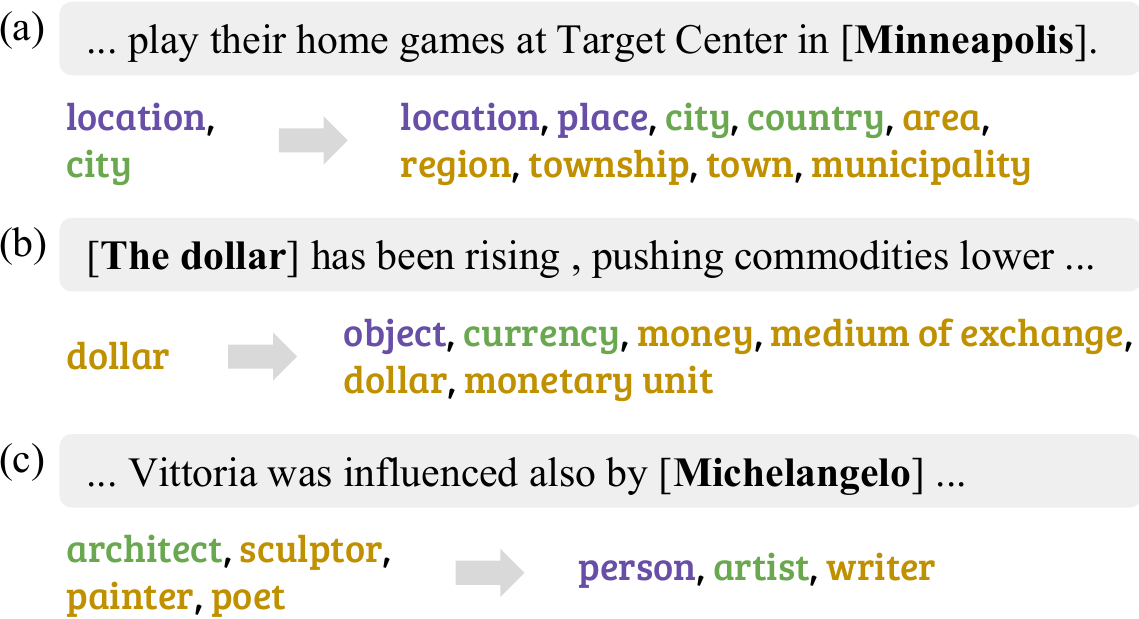}
    \caption{Examples of the noisy labels (left) and the denoised labels (right) for mentions (bold). The colors correspond to type classes: {\it general} (purple), {\it fine-grained} (green), and {\it ultra-fine} (yellow).}
    \label{fig:relabel_ex}
\end{figure}

Figure~\ref{fig:relabel_ex} shows examples of the original noisy labels and the denoised labels produced by the relabeling function. In example (a), taken from the EL data, the original labels, {\tt \{location, city\}}, are correct, but human annotators might choose more types for the mention span, {\bf Minneapolis}. The relabeling function retains the original types about the geography and adds ultra-fine types about administrative units such as {\tt \{township, municipality\}}.
In example (b), from the HEAD data, the original label, {\tt \{dollar\}}, is not so expressive by itself since it is a name of a currency. The labeling function adds coarse types, {\tt \{object, currency\}}, as well as specific types such as {\tt \{medium of exchange, monetary unit\}}. In another EL example (c), the relabeling function tries to add coarse and fine types but struggles to assign multiple diverse {\it ultra-fine} types to the mention span {\bf Michelangelo}, possibly because some of these types rarely cooccur (\texttt{painter} and \texttt{poet}).

\renewcommand{\arraystretch}{1}
\begin{table}[t]
	\centering
	\small
	\setlength{\tabcolsep}{4pt}
	\begin{tabular}{l c c c c c c c}
		\toprule
		\multicolumn{1}{c}{} & \multicolumn{2}{c}{General} & \multicolumn{2}{c}{Fine} & \multicolumn{2}{c}{Ultra-Fine} \\
	    \cmidrule(r){2-3}  \cmidrule(r){4-5} \cmidrule(r){6-7}
		\multicolumn{1}{c}{Data}
		 & Add & Del & Add & Del  & Add & Del & Filter (\%) \\
		\midrule
		EL & 0.87 & 0.01 & 0.36 & 0.17 & 2.03 & 0.12 & 9.4\\
		HEAD  & 1.18 & 0.00 & 0.51 & 0.01 & 1.15 & 0.16 & 10.0 \\
		\bottomrule 
	\end{tabular}
	\caption{The average number of types added or deleted by the relabeling function per example. The right-most column shows that the rate of examples discarded by the filtering function.} \label{tab:relabel}
\end{table}


\section{Related Work}

Past work on denoising data for entity typing has used multi-instance multi-label learning \cite{Yaghoobzadeh15, Yaghoobzadeh17, Murty18}. One view of these approaches is that they delete noisily-introduced labels, but they cannot add them, or filter bad examples. Other work focuses on learning type embeddings \cite{Yogatama15,Xiang_Ren_16a, Xiang_Ren_16b}; our approach goes beyond this in treating the label set in a structured way. The label set of \citet{Eunsol_Choi_18} is distinct in not being explicitly hierarchical, making past hierarchical approaches difficult to apply.


Denoising techniques for distant supervision have been applied extensively to relation extraction. Here, multi-instance learning and probabilistic graphical modeling approaches have been used \cite{Sebastian_Riedel10, Raphael_Hoffmann11, Mihai_Surdeanu12, Shingo_Takamatsu_12} as well as deep models \cite{Yankai_Lin16, Xiaocheng_Feng17, Bingfeng_Luo17, Kai_Lei18, Xu_Han18}, though these often focus on incorporating signals from other sources as opposed to manually labeled data.


\section{Conclusion}

In  this  work,  we investigated the problem of denoising distant data for entity typing tasks. We trained a filtering function that discards examples from the distantly labeled data that are wholly unusable and a relabeling function that repairs noisy labels for the retained examples. When distant data is processed with our best denoising model, our final trained model achieves state-of-the-art performance on an ultra-fine entity typing task.

\section*{Acknowledgments}

This work was partially supported by NSF Grant IIS-1814522, NSF Grant SHF-1762299, a Bloomberg Data Science Grant, and an equipment grant from NVIDIA. The authors acknowledge the Texas Advanced Computing Center (TACC) at The University of Texas at Austin for providing HPC resources used to conduct this research. Results presented in this paper were obtained using the Chameleon testbed supported by the National Science Foundation. Thanks as well to the anonymous reviewers for their thoughtful comments, members of the UT TAUR lab and Pengxiang Cheng for helpful discussion, and Eunsol Choi for providing the full datasets and useful resources.

\bibliography{naaclhlt2019}

\begin{thebibliography}{43}
\expandafter\ifx\csname natexlab\endcsname\relax\def\natexlab#1{#1}\fi

\bibitem[{Choi et~al.(2018)Choi, Levy, Choi, and Zettlemoyer}]{Eunsol_Choi_18}
Eunsol Choi, Omer Levy, Yejin Choi, and Luke Zettlemoyer. 2018.
\newblock {Ultra-Fine Entity Typing}.
\newblock In \emph{Proceedings of the 56th Annual Meeting of the Association
  for Computational Linguistics}.

\bibitem[{Collobert et~al.(2011)Collobert, Weston, Bottou, Karlen, Kavukcuoglu,
  and Kuksa}]{nlpfromscratch11}
Ronan Collobert, Jason Weston, L{\'{e}}on Bottou, Michael Karlen, Koray
  Kavukcuoglu, and Pavel~P. Kuksa. 2011.
\newblock \href {http://dl.acm.org/citation.cfm?id=2078186} {{Natural Language
  Processing (Almost) from Scratch}}.
\newblock \emph{Journal of Machine Learning Research}, 12:2493--2537.

\bibitem[{Del~Corro et~al.(2015)Del~Corro, Abujabal, Gemulla, and
  Weikum}]{delcorro-EtAl:2015:EMNLP}
Luciano Del~Corro, Abdalghani Abujabal, Rainer Gemulla, and Gerhard Weikum.
  2015.
\newblock {FINET: Context-Aware Fine-Grained Named Entity Typing}.
\newblock In \emph{Proceedings of the 2015 Conference on Empirical Methods in
  Natural Language Processing}.

\bibitem[{Devlin et~al.(2018)Devlin, Chang, Lee, and Toutanova}]{BERT18}
Jacob Devlin, Ming{-}Wei Chang, Kenton Lee, and Kristina Toutanova. 2018.
\newblock \href {http://arxiv.org/abs/1810.04805} {{{BERT:} Pre-training of
  Deep Bidirectional Transformers for Language Understanding}}.
\newblock \emph{CoRR}, abs/1810.04805.

\bibitem[{Feng et~al.(2017)Feng, Guo, Qin, Liu, and Liu}]{Xiaocheng_Feng17}
Xiaocheng Feng, Jiang Guo, Bing Qin, Ting Liu, and Yongjie Liu. 2017.
\newblock {Effective Deep Memory Networks for Distant Supervised Relation
  Extraction}.
\newblock In \emph{Proceedings of the Twenty-Sixth International Joint
  Conference on Artificial Intelligence, {IJCAI-17}}.

\bibitem[{Gillick et~al.(2014)Gillick, Lazic, Ganchev, Kirchner, and
  Huynh}]{Dan_Gillick_14}
Dan Gillick, Nevena Lazic, Kuzman Ganchev, Jesse Kirchner, and David Huynh.
  2014.
\newblock \href {http://arxiv.org/abs/1412.1820} {{Context-Dependent
  Fine-Grained Entity Type Tagging}}.
\newblock \emph{CoRR}, abs/1412.1820.

\bibitem[{Graves and Schmidhuber(2005)}]{bilstm}
Alex Graves and J{\"{u}}rgen Schmidhuber. 2005.
\newblock \href {https://doi.org/10.1016/j.neunet.2005.06.042} {{Framewise
  phoneme classification with bidirectional {LSTM} and other neural network
  architectures}}.
\newblock \emph{Neural Networks}, 18(5-6):602--610.

\bibitem[{Han et~al.(2018)Han, Liu, and Sun}]{Xu_Han18}
Xu~Han, Zhiyuan Liu, and Maosong Sun. 2018.
\newblock \href {http://arxiv.org/abs/1805.10959} {{Denoising Distant
  Supervision for Relation Extraction via Instance-Level Adversarial
  Training}}.
\newblock \emph{CoRR}, abs/1805.10959.

\bibitem[{Hancock et~al.(2018)Hancock, Varma, Wang, Bringmann, Liang, and
  R\'{e}}]{hancock2018}
Braden Hancock, Paroma Varma, Stephanie Wang, Martin Bringmann, Percy Liang,
  and Christopher R\'{e}. 2018.
\newblock {Training Classifiers with Natural Language Explanations}.
\newblock In \emph{Proceedings of the 56th Annual Meeting of the Association
  for Computational Linguistics}.

\bibitem[{He et~al.(2016)He, Zhang, Ren, and Sun}]{resnet16}
Kaiming He, Xiangyu Zhang, Shaoqing Ren, and Jian Sun. 2016.
\newblock {Deep Residual Learning for Image Recognition}.
\newblock In \emph{2016 {IEEE} Conference on Computer Vision and Pattern
  Recognition, {CVPR}}.

\bibitem[{Hochreiter and Schmidhuber(1997)}]{lstm}
Sepp Hochreiter and J{\"u}rgen Schmidhuber. 1997.
\newblock {Long short-term memory}.
\newblock \emph{Neural computation}, 9(8):1735--1780.

\bibitem[{Hoffmann et~al.(2011)Hoffmann, Zhang, Ling, Zettlemoyer, and
  Weld}]{Raphael_Hoffmann11}
Raphael Hoffmann, Congle Zhang, Xiao Ling, Luke~S. Zettlemoyer, and Daniel~S.
  Weld. 2011.
\newblock {Knowledge-Based Weak Supervision for Information Extraction of
  Overlapping Relations}.
\newblock In \emph{Proceedings of the 49th Annual Meeting of the Association
  for Computational Linguistics: Human Language Technologies}.

\bibitem[{Honnibal and Johnson(2015)}]{spacy_15}
Matthew Honnibal and Mark Johnson. 2015.
\newblock {An Improved Non-monotonic Transition System for Dependency Parsing}.
\newblock In \emph{Proceedings of the 2015 Conference on Empirical Methods in
  Natural Language Processing}.

\bibitem[{Hovy et~al.(2006)Hovy, Marcus, Palmer, Ramshaw, and
  Weischedel}]{HovyEtAl2006}
Eduard Hovy, Mitchell Marcus, Martha Palmer, Lance Ramshaw, and Ralph
  Weischedel. 2006.
\newblock {OntoNotes: The 90\% Solution}.
\newblock In \emph{Proceedings of the 2006 Conference of the North American
  Chapter of the Association for Computational Linguistics: Human Language
  Technologies}.

\bibitem[{Kingma and Ba(2014)}]{Kingma_14}
Diederik~P. Kingma and Jimmy Ba. 2014.
\newblock \href {http://arxiv.org/abs/1412.6980} {{Adam: {A} Method for
  Stochastic Optimization}}.
\newblock \emph{CoRR}, abs/1412.6980.

\bibitem[{Lee et~al.(2017)Lee, He, Lewis, and Zettlemoyer}]{Kenton_Lee17}
Kenton Lee, Luheng He, Mike Lewis, and Luke Zettlemoyer. 2017.
\newblock {End-to-end Neural Coreference Resolution}.
\newblock In \emph{Proceedings of the 2017 Conference on Empirical Methods in
  Natural Language Processing}.

\bibitem[{Lei et~al.(2018)Lei, Chen, Li, Du, Yang, Fan, and Shen}]{Kai_Lei18}
Kai Lei, Daoyuan Chen, Yaliang Li, Nan Du, Min Yang, Wei Fan, and Ying Shen.
  2018.
\newblock {Cooperative Denoising for Distantly Supervised Relation Extraction}.
\newblock In \emph{Proceedings of the 27th International Conference on
  Computational Linguistics}.

\bibitem[{Lin et~al.(2016)Lin, Shen, Liu, Luan, and Sun}]{Yankai_Lin16}
Yankai Lin, Shiqi Shen, Zhiyuan Liu, Huanbo Luan, and Maosong Sun. 2016.
\newblock {Neural Relation Extraction with Selective Attention over Instances}.
\newblock In \emph{Proceedings of the 54th Annual Meeting of the Association
  for Computational Linguistics}.

\bibitem[{Luo et~al.(2017)Luo, Feng, Wang, Zhu, Huang, Yan, and
  Zhao}]{Bingfeng_Luo17}
Bingfeng Luo, Yansong Feng, Zheng Wang, Zhanxing Zhu, Songfang Huang, Rui Yan,
  and Dongyan Zhao. 2017.
\newblock {Learning with Noise: Enhance Distantly Supervised Relation
  Extraction with Dynamic Transition Matrix}.
\newblock In \emph{Proceedings of the 55th Annual Meeting of the Association
  for Computational Linguistics}.

\bibitem[{Miller(1995)}]{wordnet_1995}
George~A. Miller. 1995.
\newblock \href {https://doi.org/10.1145/219717.219748} {{WordNet: A Lexical
  Database for English}}.
\newblock \emph{Commun. ACM}, 38(11):39--41.

\bibitem[{Milne and Witten(2008)}]{MilneWitten2008}
David Milne and Ian~H. Witten. 2008.
\newblock {Learning to Link with Wikipedia}.
\newblock In \emph{Proceedings of the 17th ACM Conference on Information and
  Knowledge Management}.

\bibitem[{Mintz et~al.(2009)Mintz, Bills, Snow, and Jurafsky}]{Mike_Mintz_09}
Mike Mintz, Steven Bills, Rion Snow, and Daniel Jurafsky. 2009.
\newblock {Distant supervision for relation extraction without labeled data}.
\newblock In \emph{Proceedings of the Joint Conference of the 47th Annual
  Meeting of the ACL and the 4th International Joint Conference on Natural
  Language Processing of the AFNLP}.

\bibitem[{Murty et~al.(2018)Murty, Verga, Vilnis, Radovanovic, and
  McCallum}]{Murty18}
Shikhar Murty, Patrick Verga, Luke Vilnis, Irena Radovanovic, and Andrew
  McCallum. 2018.
\newblock {Hierarchical Losses and New Resources for Fine-grained Entity Typing
  and Linking}.
\newblock In \emph{Proceedings of the 56th Annual Meeting of the Association
  for Computational Linguistics}.

\bibitem[{Pennington et~al.(2014)Pennington, Socher, and Manning}]{glove_14}
Jeffrey Pennington, Richard Socher, and Christopher~D. Manning. 2014.
\newblock {Glove: Global Vectors for Word Representation}.
\newblock In \emph{Proceedings of the 2014 Conference on Empirical Methods in
  Natural Language Processing}.

\bibitem[{Peters et~al.(2018)Peters, Neumann, Iyyer, Gardner, Clark, Lee, and
  Zettlemoyer}]{ELMO_18}
Matthew~E. Peters, Mark Neumann, Mohit Iyyer, Matt Gardner, Christopher Clark,
  Kenton Lee, and Luke Zettlemoyer. 2018.
\newblock {Deep Contextualized Word Representations}.
\newblock In \emph{Proceedings of the 2018 Conference of the North American
  Chapter of the Association for Computational Linguistics: Human Language
  Technologies}.

\bibitem[{Rabinovich and Klein(2017)}]{rabinovich-klein:2017:Short}
Maxim Rabinovich and Dan Klein. 2017.
\newblock {Fine-Grained Entity Typing with High-Multiplicity Assignments}.
\newblock In \emph{Proceedings of the 55th Annual Meeting of the Association
  for Computational Linguistics}.

\bibitem[{Ratner et~al.(2018{\natexlab{a}})Ratner, Bach, Ehrenberg, Fries, Wu,
  and R\'{e}}]{ratner-snorkel}
Alexander Ratner, Stephen~H. Bach, Henry Ehrenberg, Jason Fries, Sen Wu, and
  Christopher R\'{e}. 2018{\natexlab{a}}.
\newblock {Snorkel: Rapid Training Data Creation with Weak Supervision}.
\newblock In \emph{Proceedings of VLDB}.

\bibitem[{Ratner et~al.(2018{\natexlab{b}})Ratner, Hancock, Dunnmon, Sala,
  Pandey, and R{\'{e}}}]{Alexander_Ratner_18}
Alexander Ratner, Braden Hancock, Jared Dunnmon, Frederic Sala, Shreyash
  Pandey, and Christopher R{\'{e}}. 2018{\natexlab{b}}.
\newblock \href {http://arxiv.org/abs/1810.02840} {{Training Complex Models
  with Multi-Task Weak Supervision}}.
\newblock \emph{CoRR}, abs/1810.02840.

\bibitem[{Ratner et~al.(2016)Ratner, Sa, Wu, Selsam, , and
  R\'{e}}]{ratner16-data-programming}
Alexander Ratner, Chris~De Sa, Sen Wu, Daniel Selsam, , and Christopher R\'{e}.
  2016.
\newblock {Data Programming: Creating Large Training Sets, Quickly}.
\newblock In \emph{Proceedings of NeurIPS}.

\bibitem[{Ren et~al.(2016{\natexlab{a}})Ren, He, Qu, Huang, Ji, and
  Han}]{Xiang_Ren_16a}
Xiang Ren, Wenqi He, Meng Qu, Lifu Huang, Heng Ji, and Jiawei Han.
  2016{\natexlab{a}}.
\newblock {{AFET:} Automatic Fine-Grained Entity Typing by Hierarchical
  Partial-Label Embedding}.
\newblock In \emph{Proceedings of the 2016 Conference on Empirical Methods in
  Natural Language Processing}.

\bibitem[{Ren et~al.(2016{\natexlab{b}})Ren, He, Qu, Voss, Ji, and
  Han}]{Xiang_Ren_16b}
Xiang Ren, Wenqi He, Meng Qu, Clare~R. Voss, Heng Ji, and Jiawei Han.
  2016{\natexlab{b}}.
\newblock {Label Noise Reduction in Entity Typing by Heterogeneous
  Partial-Label Embedding}.
\newblock In \emph{Proceedings of the 22nd ACM SIGKDD International Conference
  on Knowledge Discovery and Data Mining}.

\bibitem[{Riedel et~al.(2010)Riedel, Yao, and McCallum}]{Sebastian_Riedel10}
Sebastian Riedel, Limin Yao, and Andrew McCallum. 2010.
\newblock {Modeling Relations and Their Mentions without Labeled Text}.
\newblock In \emph{Machine Learning and Knowledge Discovery in Databases}.

\bibitem[{Shimaoka et~al.(2017)Shimaoka, Stenetorp, Inui, and
  Riedel}]{Sonse_Shimaoka_17}
Sonse Shimaoka, Pontus Stenetorp, Kentaro Inui, and Sebastian Riedel. 2017.
\newblock {Neural Architectures for Fine-grained Entity Type Classification}.
\newblock In \emph{Proceedings of the 15th Conference of the European Chapter
  of the Association for Computational Linguistics}.

\bibitem[{Srivastava et~al.(2015)Srivastava, Greff, and
  Schmidhuber}]{highway15}
Rupesh~Kumar Srivastava, Klaus Greff, and J{\"{u}}rgen Schmidhuber. 2015.
\newblock \href {http://arxiv.org/abs/1505.00387} {{Highway Networks}}.
\newblock \emph{CoRR}, abs/1505.00387.

\bibitem[{Surdeanu et~al.(2012)Surdeanu, Tibshirani, Nallapati, and
  Manning}]{Mihai_Surdeanu12}
Mihai Surdeanu, Julie Tibshirani, Ramesh Nallapati, and Christopher~D. Manning.
  2012.
\newblock {Multi-instance Multi-label Learning for Relation Extraction}.
\newblock In \emph{Proceedings of the 2012 Joint Conference on Empirical
  Methods in Natural Language Processing and Computational Natural Language
  Learning}.

\bibitem[{Takamatsu et~al.(2012)Takamatsu, Sato, and
  Nakagawa}]{Shingo_Takamatsu_12}
Shingo Takamatsu, Issei Sato, and Hiroshi Nakagawa. 2012.
\newblock {Reducing Wrong Labels in Distant Supervision for Relation
  Extraction}.
\newblock In \emph{Proceedings of the 50th Annual Meeting of the Association
  for Computational Linguistics}.

\bibitem[{Vincent et~al.(2008)Vincent, Larochelle, Bengio, and
  Manzagol}]{Pascal_Vincent_08}
Pascal Vincent, Hugo Larochelle, Yoshua Bengio, and Pierre{-}Antoine Manzagol.
  2008.
\newblock {Extracting and composing robust features with denoising
  autoencoders}.
\newblock In \emph{Proceedings of the 25th International Conference on Machine
  Learning}.

\bibitem[{Wang and Poon(2018)}]{Hai_Wang_18}
Hai Wang and Hoifung Poon. 2018.
\newblock {Deep Probabilistic Logic: {A} Unifying Framework for Indirect
  Supervision}.
\newblock In \emph{Proceedings of the 2018 Conference on Empirical Methods in
  Natural Language Processing}.

\bibitem[{Xiong et~al.(2019)Xiong, Wu, Lei, Yu, Chang, Guo, and
  Wang}]{Wenhan_Xiong_2019}
Wenhan Xiong, Jiawei Wu, Deren Lei, Mo~Yu, Shiyu Chang, Xiaoxiao Guo, and
  William~Yang Wang. 2019.
\newblock {Imposing Label-Relational Inductive Bias for Extremely Fine-Grained
  Entity Typing}.
\newblock In \emph{Proceedings of the 2019 Conference of the North American
  Chapter of the Association for Computational Linguistics: Human Language
  Technologies}.

\bibitem[{Yaghoobzadeh and Sch{\"{u}}tze(2015)}]{Yaghoobzadeh15}
Yadollah Yaghoobzadeh and Hinrich Sch{\"{u}}tze. 2015.
\newblock {Corpus-level Fine-grained Entity Typing Using Contextual
  Information}.
\newblock In \emph{Proceedings of the 2015 Conference on Empirical Methods in
  Natural Language Processing}.

\bibitem[{Yaghoobzadeh and Sch{\"{u}}tze(2017)}]{Yaghoobzadeh17}
Yadollah Yaghoobzadeh and Hinrich Sch{\"{u}}tze. 2017.
\newblock {Multi-level Representations for Fine-Grained Typing of Knowledge
  Base Entities}.
\newblock In \emph{Proceedings of the 15th Conference of the European Chapter
  of the Association for Computational Linguistics}.

\bibitem[{Yogatama et~al.(2015)Yogatama, Gillick, and Lazic}]{Yogatama15}
Dani Yogatama, Daniel Gillick, and Nevena Lazic. 2015.
\newblock {Embedding Methods for Fine Grained Entity Type Classification}.
\newblock In \emph{Proceedings of the 53rd Annual Meeting of the Association
  for Computational Linguistics and the 7th International Joint Conference on
  Natural Language Processing}.

\bibitem[{Zhou and Li(2005)}]{tri05}
Zhi{-}Hua Zhou and Ming Li. 2005.
\newblock \href {https://doi.org/10.1109/TKDE.2005.186} {{Tri-Training:
  Exploiting Unlabeled Data Using Three Classifiers}}.
\newblock \emph{{IEEE} Trans. Knowl. Data Eng.}, 17(11):1529--1541.

\end{thebibliography}
\bibliographystyle{acl_natbib}

\end{document}